# Towards a Linguistic Evaluation of Narratives: A Quantitative Stylistic Framework*

Alessandro Maisto[1,*,†]

[1]*Università degli Studi di Salerno, Via Giovanni Paolo II, 132, 84084 Fisciano (SA), Italia*

**Abstract**

The evaluation of narrative quality remains a complex challenge, as it involves subjective factors such as plot, character development, and emotional impact. This work proposes a quantitative approach to narrative assessment by focusing on the linguistic dimension as a primary indicator of quality. The paper presents a methodology for the automatic evaluation of narrative based on the extraction of a comprehensive set of 33 quantitative linguistic features categorized into lexical, syntactic, and semantic groups. To test the model, an experiment was conducted on a specialized corpus of 23 books, including canonical masterpieces and self-published works. Through a similarity matrix, the system successfully clustered the narratives, distinguishing almost perfectly between professionally edited and self-published texts. Furthermore, the methodology was validated against a human-annotated dataset; it significantly outperforms traditional story-level evaluation metrics, demonstrating the effectiveness of quantitative linguistic features in assessing narrative quality.

## 1. Introduction

The success of a narrative can be attributed to numerous factors. While the plot is undoubtedly a crucial element of a good story, other factors such as character plausibility, quality of dialogue, pacing, emotions and world-building also play vital roles [1, 2, 3]. However, a written story that excels in all these areas cannot be considered a high-quality narrative—nor can it reach a wide audience—if its linguistic dimension is poor. A lack of coherence, cohesion, syntactic organization, or semantic depth can transform a potentially good narrative into an inevitable failure.

Assessing the linguistic boundaries of a narrative is no easy task; however, it may be simpler than evaluating non-linguistic features. In fact, although a linguistic evaluation covers various features of different natures (morphological, semantic, syntactic, and pragmatic), language must adhere to schematic rules that, in many cases, can be automatically evaluated in a quantitative manner.

The aim of this paper is to test the ability of diverse linguistic stylistic features to classify and evaluate a written narration in an unsupervised fashion. To achieve this, I conducted an experiment in which 33 quantitative linguistic features were extracted from a corpus of 23 books to create representative numerical vectors. Each vector was compared using a correlation measure to extract similarity values. Subsequently, using a network classification algorithm known as "modularity class," the data was clustered for unsupervised classification. As previously mentioned, the extraction of linguistic features is a minor obstacle, as it is facilitated by efficient parsing algorithms—such as the CoreNLP package—or modern semantic models like BERT.

Furthermore, the 23 classified and vectorized books served as the foundational "gold standard" for a new evaluation metric. This methodology was applied to a larger corpus comprising 1056 human-evaluated, human-written and AI-generated texts from the HANNA dataset. By establishing this framework, the new metric could be directly compared against traditional evaluation standards by calculating the correlation, confirming that linguistic features are highly effective in assessing narrative quality.

However, selecting the corpus represents a significant challenge, as almost all published books hold relevance for specific groups of readers, regardless of sales figures. To build a representative corpus, I based my research on four categories: on the one hand, I selected examples of effective narratives universally recognized as masterpieces. To do this, I first consulted the list of the most-read English books and then selected titles awarded by prestigious institutions (e.g., the Pulitzer Prize, the Nobel Prize). On the other hand, I required books that are not universally acclaimed or are dismissed by critics. For this category, I focused on two types of texts: mass-market romance best-sellers and romance fan-fiction. I am aware that this selection is problematic from different points of view: the quality of a book is not directly related to whether it has received an award or passed a professional revision. In addition, the style of each book, although they belong to the same class, could be so different that these differences can interfere with the final linguistic classification. Nevertheless, the first experiment will be used as a preliminary test to determine to what extent these 23 books can work as a gold standard for a more general evaluation metric.

The results of the automatic classification indicate that the approach taken in this experiment warrants further investigation. The findings confirm that the selected linguistic features perform well in classifying the quality of narratives. Therefore, a larger study involving more books and additional features is necessary.

Nevertheless, the application of this methodology—utilizing the 23 classified books as a gold standard—led to the development of a new evaluation metric that achieved successful results in automatic assessment. When validated against the human-annotated dataset, this metric outperformed almost all traditional evaluation methods.

The paper is structured as follows: Section 2 briefly presents the state of the art regarding linguistic evaluation metrics for narratives; Section 3 describes the methodology adopted; Section 4 illustrates the experimental results; and Section 5 presents my conclusions and proposals for future work.

## 2. Related Works

The evaluation of narratives can be divided into two distinct and equally challenging tasks: human and automatic evaluation. The former, human evaluation, is in fact as difficult as the latter [4, 1, 2]. Nevertheless, human evaluation must be considered the "gold standard", even though it is very expensive and time-consuming. As underlined by Chhun et al. [5], human criteria include concepts such as coherence [6, 7], overall quality [8], narrative flow [9], and emotion [3].

Coherence is defined as the connection between the different sentences of a single story. In many automatically generated narratives, the lack of connectives [10, 11] reduces text coherence, leading to comprehension problems for human readers. Overall quality regards the human judgment of the story as a whole, including, in some cases, how well the story matches the prompt used in AI-generated texts [12]. Narrative flow relates to the order in which information is presented and the logical and temporal consequentiality of events. In Rashkin et al. [9], it is associated with repeated information between paragraphs, transitions, and topical coherence. Emotion can refer to empathy for the characters, external emotional responses, or humor, and is closely related to the "interestingness" of the story [3].

Concerning automatic evaluation, most studies rely on Machine Translation Evaluation (MTE) metrics such as BLEU [13] or ROUGE [14]. As in MTE, these systems can be *reference-based*, if there is a reference text from which the distance in terms of lexicon or syntax is calculated, or *reference-free*, which generally rely only on the generated text or the prompt [15, 16]. Recently, many models have been based on embeddings [17] or pre-trained language models [18, 19, 20]. For the latter, the most important factor is prompting [21], since the evaluation prompt can include a simple rating, a rating with explanation, with the addition of guidelines, but also with the human reference stories [4].

All the papers cited above employ narrative evaluation to measure the quality of automatically generated stories. However, it is fundamental to understand how to evaluate literature to define realistic criteria for narrative assessment. Dickman [2] identifies four elements that characterize "every successful story." He uses the acronym PHAT to represent his model: Passion, Hero, Antagonist, and Transformation. Nevertheless, PHAT criteria are difficult to measure computationally.

McCabe and Peterson [1] proposed three ways to analyze the structure of a story: the first approaches a story as an ordered set of episodes to emphasize goals and their achievement; the second is related to affective information, focusing on emotional peaks or crisis events; the third approach emphasizes linguistic complexity by studying the connections among propositions.

More recently, van Dalen-Oskam [22] performed a large-scale semi-automatic evaluation of literary narratives based on the extraction of frequent words. Frequent words have been employed to perform a cluster analysis in which texts with similar word frequencies are statistically linked in pairs.

Many studies employ the *Tool for the Automatic Analysis of Cohesion* (TAACO) [23, 24] to evaluate coherence in essays or novels [25], but recently, the use of LLMs has also become common in this field [26].

## 3. Methodology

In this work, a classification test over a set of measures for the linguistic assessment of narratives proceed a larger evaluation experiment. To verify the efficiency of the selected linguistic features on real-world texts, our methodology follows these steps:

1. Define the set of linguistic features previously employed in literature to evaluate natural language narratives;
2. Identify a corpus of narratives with established reputations—either universally acclaimed or dismissed by critics and the public;
3. Convert these narratives into normalized, comparable numerical vectors to generate a similarity network and classify the texts;
4. Extend the analysis to a human-rated database to validate the methodology on external data.

The following subsections briefly detail each stage of this methodology, while the subsequent section presents the experimental results.

### 3.1. Linguistic Features

As discussed in Section 2, various metrics exist for narrative evaluation. This study moves beyond traditional methods—such as MTE metrics, embeddings, or LLMs—to extract a comprehensive set of quantitative linguistic features directly from the texts. To achieve this, we utilize two primary instruments: the Stanford Parser [27] for dependency parsing and BERT [28] for semantic metrics.

The 33 identified linguistic features are categorized into three groups: Lexical, Syntactic, and Semantic (see table 1).

#### 3.1.1. Lexical Features

Lexical features are straightforward to extract as they rely only on text tokenization. Our set incorporates several metrics used by Maisto [10] to differentiate between oral and written texts, supplemented by measures inspired by the TAACO model. The specific features employed are:

- **D-Textual-Value**: inspired by D-Value [29], this is a measure of lexical diversity similar to the Type-Token Ratio (TTR) based on the number of word of each paragraph and the number of polysyllabic words.
- **Lexical Range**: calculation of word frequencies across three tiers: the 1,000 (**LR1**) and 2,000 (**LR2**) most frequent words from the "General Service List" [30], and the "Academic Word List" (**LR3**) [31];

**Table 1**
Summary of the Classification

| Lexical (20) | Syntactic (10) | Semantic (3) |
|---|---|---|
| D-Text-Value | Relative Ratio | |
| LR1, LR2, LR3 | Present tense Ratio | Average Semantic Overlap |
| Concreteness | Past tense Ratio | General Figurative Index |
| Noun/Pronoun Ratio | Participle Tense Ratio | Sentences Figurative Index |
| Deictic/Articles Ratio | Modifier per Noun | |
| Freq. of Definite articles | Average Graph Depth | |
| Freq. of Attributive Adjectives | Temporal Stability | |
| Freq. of Emphatic Particles | Hypotactic Depth | |
| Freq. of Demonstrative | Verb Density | |
| First Person Ratio | | |
| Hapax Legomena | | |
| Pos and Neg Additive, Logical, Temporal, Causal connectives | | |

- **Concreteness**: based on the MRC Psycholinguistic Database [32], assigning values (100–700) to gauge the "perceptibility" of the vocabulary;
- **POS Ratio**: proportions of Part-of-Speech tags, including **noun/pronoun ratio**, **deictic/article ratio**, **first person pronouns/pronouns**, and the **frequency of determiners**, **adjectives**, and **emphatic particles** per sentence;
- **Lexical Overlap**: frequency of content-word repetitions across two or three adjacent sentences;
- **Hapax Legomena**: the ratio of unique words (frequency of 1) to the total word count;
- **Connective Extraction**: a count of **positive** and **negative additive**, **causal**, **temporal**, and **logical** connectives used as a proxy for text **cohesion**.

These features were selected for their ability to quantify text quality. While TAACO typically utilizes TTR, we chose D-Textual-Value to account for the significantly varying lengths within our corpus. Additionally, Lexical Range serves as an established metric for readability [33], while Concreteness reflects the balance between action and introspection. Finally, connectives are vital for signaling internal relationships [34] and provide key insights into overall writing quality [35].

### 3.1.2. Syntactical Features

The selection of syntactic features is also partially based on Maisto [10], supplemented by three novel measures focusing on verb tense and *consecutio temporum*. These metrics are derived from dependency parsing performed using the CoreNLP Java package [27]. The features are as follows:

- **Relative/Subordinate Ratio**: the proportion of relative and subordinate clauses.
- **Verb Tense Ratios**: proportions of **present**, **participle**, and **past** tense verbs.
- **Modifier-per-Noun Ratio**: the density of modifiers relative to nouns.
- **Average Graph Depth**: the maximum number of nodes separating the root from any other node in the dependency tree.
- **Verb Density**: the percentage of verbs on tokens per sentence.
- **Temporal Stability**: a metric measuring tense consistency between adjacent sentences; cohesive texts typically maintain temporal stability across proximal sentences.
- **Hypotactic Depth**: a nested graph representing verb tenses in a parenthetical dependency system. Levels range from 0 (main clause only) to 2 or more (multiple levels of subordination).

While the extraction of the initial metrics are simply based on the POS and dependency tags, the final two require complex preprocessing. First, complex verb tenses are unified; auxiliary verbs are not treated as independent units but are merged with their respective heads.

To calculate *Hypotactic Depth*, the model generates a simplified schema where the ROOT verb is marked with asterisks and subordinated verbs are enclosed in parentheses. For example, "I hoped that she wanted to go" is represented as *Past*(Past(Inf)). The system then calculates the average number of nested parentheses per sentence.

Regarding *Temporal Stability*, the system extracts the tense of the ROOT verbs from every sentence's dependency graph. The dominant tense of the document is identified, and stability is calculated as the ratio of dominant verbs (including compatible tenses like Present-Future-Progressive) to the total number of verbs.

### 3.1.3. Semantic Features

The final group of metrics consists of semantic features. We utilized the pre-trained models "bert-base-uncased" and "all-MiniLM-L6-v2" to calculate the following indices:

- **Average Semantic Overlap**: the semantic similarity between adjacent sentences.
- **Textual Figurative Indices**: a measure of the semantic predictability of operator-argument relations within the text.

The first metric leverages "all-MiniLM-L6-v2" to generate sentence embeddings, comparing each sentence to the one following it. The final index for a given text is the arithmetic mean of these sentence-to-sentence similarities.

The *Textual Figurative Indices* use the BERT model to predict the 20 most probable words for a specific position (Masked Language Modeling) and calculate the average semantic distance between the actual word and these top 20 candidates. The operator-argument relations analyzed include:

- Subject-verb (e.g., "The wind howled");
- Subject-nominal/adjectival head (e.g., "Love is a battlefield");
- Verb-object (e.g., "Harvest the ideas");
- Passive subject-verb (e.g., "The city was wrapped in fog").

For instance, in the sentence "The Dursleys couldn't bear the idea of magic," the system predicts the 20 most probable words to replace the subject ("[MASK] couldn't bear...") and the verb ("The Dursleys couldn't [MASK]..."). Low average similarity between the actual elements and the predicted candidates indicates the presence of figurative language. In this process, proper nouns (e.g., "Dursley") are replaced by pronouns to avoid bias.

In cases like "Time is a thief," where BERT might actually predict the metaphor due to the frequency of the sequence, a secondary control routine performs a direct cosine similarity check between the embeddings (e.g., "Time" and "thief") in a neutral context. If the similarity is low, the sequence is tagged as figurative.

The final *Textual Figurative Indices* are calculated by dividing the total number of figurative relations by both the number of sentences (TFI/Sent) and the number of tokens, then multiplying by 1,000 (TFI/1000).

### 3.2. Corpus Selection

To quantify the presence of the selected features and investigate the extent to which they indicate a successful narrative, a specialized corpus is required. While selecting novels that represent high-quality narratives is undoubtedly a challenging task, identifying what we might term "substandard" or "marginal" narratives is even more complex. By definition, most published books undergo extensive linguistic revision by professional editors. Nevertheless, certain works distinguish themselves through exceptional sales figures or critical acclaim. Based on these distinctions, we categorized our corpus into four narrative types:

**Table 2**
Composition of the Corpus

| HQ Best-sellers | Awarded | Standard-Quality Best-sellers | Self-published |
|---|---|---|---|
| The bridges of Madison County | To kill a mockingbird | Bared to you | Conjuring Dragons |
| The Hobbit | Beloved | Fifty shade of Grey | Arranged |
| And then there were none | The remains of the day | Twilight | A Beautiful, Terrible Love |
| The Da Vinci code | The old man and the sea | | My rejected heart |
| Harry Potter and the philosopher's stone | | | Boarding Schools, Secrets, and Jerks |
| | | | Vicious In Love |
| | | | A howl in the night |
| | | | Shut out |
| | | | When Summer Ends |
| | | | Enemies with benefits |

- **Award-winning Narratives** (Aw): best-selling works that have also received prestigious literary prizes;
- **High Quality Best-sellers** (HQ): books with universal appeal that have sold over 50 million copies;
- **Standard Quality Best-sellers** (SQ): books with high commercial success that have been largely panned by critics;
- **Self-published Works** (SP): narratives that have not undergone professional editing and for which no sales data is available.

Table 2 reports the 23 books selected and its original classification. All the selected books were originally written in English and date from the twentieth century onward. For the self-published category, the source material was obtained from the Hugging Face repository "AlekseyKorshuk/romance-books", from which ten of the largest files were selected.

The High-Quality (HQ) Best-sellers represent some of the most widely read books in English history, each with at least 50 million copies sold. For instance, the first volume of *Harry Potter* has sold 120 million copies, while *And Then There Were None* and *The Hobbit* have reached 100 million. Similarly, *The Da Vinci Code* and *The Bridges of Madison County* have sold 80 and 50 million copies, respectively.

Some awarded books also rank among the most read — such as *To Kill a Mockingbird*, with 40 million copies sold — but were categorized separately due to their critical reception. *To Kill a Mockingbird*, for example, won the Pulitzer Prize, an honor also shared by *Beloved* and *The Old Man and the Sea*. Additionally, *Midnight's Children* and *The Remains of the Day* are notable recipients of the Booker Prize.

The Standard Quality Best-sellers include books that have also sold millions of copies but are often dismissed by critics due to their themes or origins. For example, *Bared to You* was originally self-published, while *Fifty Shades of Grey* began as a *Twilight* fan-fiction.

This classification may be considered arbitrary due to the stylistic discrepancies between books within the same class, as well as the fact that the quality of a self-published work can be high despite the lack of professional post-editing. Furthermore, the distribution of volumes across categories may appear inconsistent, as the number of books varies in each class.

Nevertheless, this distribution resulted from a collection process that initially focused on ten positive and ten negative examples. Subsequently, it was decided to establish an additional class for award-winning books. The creation of this class necessitated an intermediate category, leading to the selection of three best-selling books which, for various reasons, have not received critical acclaim.

Differences in style and literary genre were considered negligible, as the linguistic features under analysis must remain as general as possible. For this reason, the initial experiment involves an unsupervised classification: if this stage generates a plausible or probable categorization with minimal errors, these features could then serve as reliable indicators of quality independently from the genre or the style of the books.

## 3.3. Similarity score and Clustering

Each selected book was analyzed using a script in Java designed ad-hoc for the experiments. The first experiment conduct to an automatic classification of the 23 books. In the second experiment, the same 23 books were used as gold-standard to generate an evaluation metric for more than 1000 texts from the HANNA dataset.

### 3.3.1. Unsupervised Classification of Books

The extraction process yielded a matrix of 33 linguistic features across 23 books. Given the significant variance in feature scales (e.g., concreteness scores averaging 300 versus decimal textual metrics), the data were normalized to ensure equal weighting in subsequent analyses. Each score was divided by its respective feature mean and multiplied by 100, expressing the values as a percentage of the mean (mean-scaling). This transformation preserves the variance across narratives while neutralizing absolute scale differences.

Subsequently, a book-by-book similarity matrix was generated using Pearson correlation coefficients between the resulting 33-dimensional vectors. To sharpen the representational power of the network and facilitate classification, we applied the transformation described by Rohde et al. [36]: negative correlations were truncated to zero, while positive coefficients were square-rooted. This process non-linearly enhances stronger similarities, producing a more distinct topological structure for the similarity network.

The matrix construction followed these steps:

1. Extraction of 33 linguistic measures for each book;
2. Normalization of scores via mean-scaling;
3. Representation of each book as a normalized 33-dimensional vector;
4. Generation of a similarity matrix using Pearson correlation;
5. Refinement of the matrix by zeroing negative values and applying a square root to positive ones.

Following the methodology of Maisto et al. [37], a network based on the similarity scores was generated. The Modularity Class algorithm [38] was then applied to the network to identify distinct clusters. This algorithm automatically detects groupings within the network in an unsupervised manner, identifying distinct clusters that correspond to different narrative classes.

Using an unsupervised classification grants that the generated classification does not influenced by the human generated classes. There is not a learning step in which the model acquires information from human generated classes and the final results is totally dependent from the linguistic features that composes the book's matrix.

### 3.3.2. Narrative Quality Evaluation

In the second experiment, the vectors of the 23 selected books were compared against the HANNA dataset [5], which comprises narratives generated by both humans and LLMs. The HANNA dataset is fully human-annotated across six dimensions: Relevance, Coherence, Empathy, Surprise, Engagement, and Complexity. Following the authors' recommendations, these dimensions were averaged to derive a composite general quality score.

The HANNA narratives were processed using the same 33 linguistic features and matrix generation workflow applied to the 23 Gold-Standard books. However, the normalization step was modified: instead of using the HANNA internal averages, the scores were normalized using the mean-scaling parameters derived from the Gold-Standard corpus. By adopting the Gold-Standard as a fixed baseline, the evaluation metrics remain independent of the HANNA dataset's composition and size, ensuring a direct and unbiased comparison between the two corpora.

Additionally, a weighting scheme was applied to the linguistic metrics in the second experiment. This was motivated by the findings of the first experiment, which indicated that certain variables introduced statistical noise when employed to evaluate narrative quality. To mitigate this, the mean values of specific metrics were scaled by a coefficient ranging from 10 to 1000 while some metrics were directly excluded from the analysis. Through iterative testing, the optimal configuration was identified by applying a scaling factor of 100 to the following 10 metrics:

- Demonstrative Ratio
- Deictic Ratio
- Relative Ratio
- Past Ratio
- Negative Additive Connectives
- Positive Causal Connectives
- Negative Causal Connectives
- Positive Temporal Connectives
- Negative Temporal Connectives
- Negative Logical Connectives

These metrics are not completely deleted, but their influence in the final score is significantly lower, determining an increasing values for both, Pearson and Kendall correlations. The following metrics, on contrary, have not been considered in the second experiment:

- First Person Pronouns Ratio
- LR1
- LR2
- LR3

To derive the final quality score, the average similarity between the Gold-Standard classes and each new text was calculated. Based on the previous classification results, three different grouping strategies were tested:

1. Original: The classes originally proposed[1];
2. Automatic: The classes automatically generated in the first experiment[2];
3. Merged: A merged model where High-Quality (HQ) Bestsellers and Awarded books were grouped, as were Standard-Quality (SQ) Bestsellers and Self-Published (SP) works[3];

A Differential Polarity Index (DPI) was subsequently developed based on combinations of these average similarities. To validate the index, we tested various formulas and evaluated their performance by comparing Pearson correlation and Kendall's Tau coefficients against human annotations.

## 4. Experiments

This section present the details of the two experiments carried out to test the identified linguistic features: Automatic Unsupervised Classification of Books and Evaluation of short narratives. For each experiment I will comment the results focusing on strengths and weaknesses of the proposed model.

### 4.1. Unsupervised Classification of Books

The similarity network was generated using Gephi [39]. The classification results are illustrated in Figure 4.1.

In Figure 4.1, each node represents a book, with colors indicating the clusters identified by the Modularity Class algorithm. Modularity is a heuristic method that extracts community structure from networks taking into account the connection weights. It has a parameter called Resolution which has been set up to 1.0 in order to not increases too much the number of generated communities. The algorithm yielded three classes.

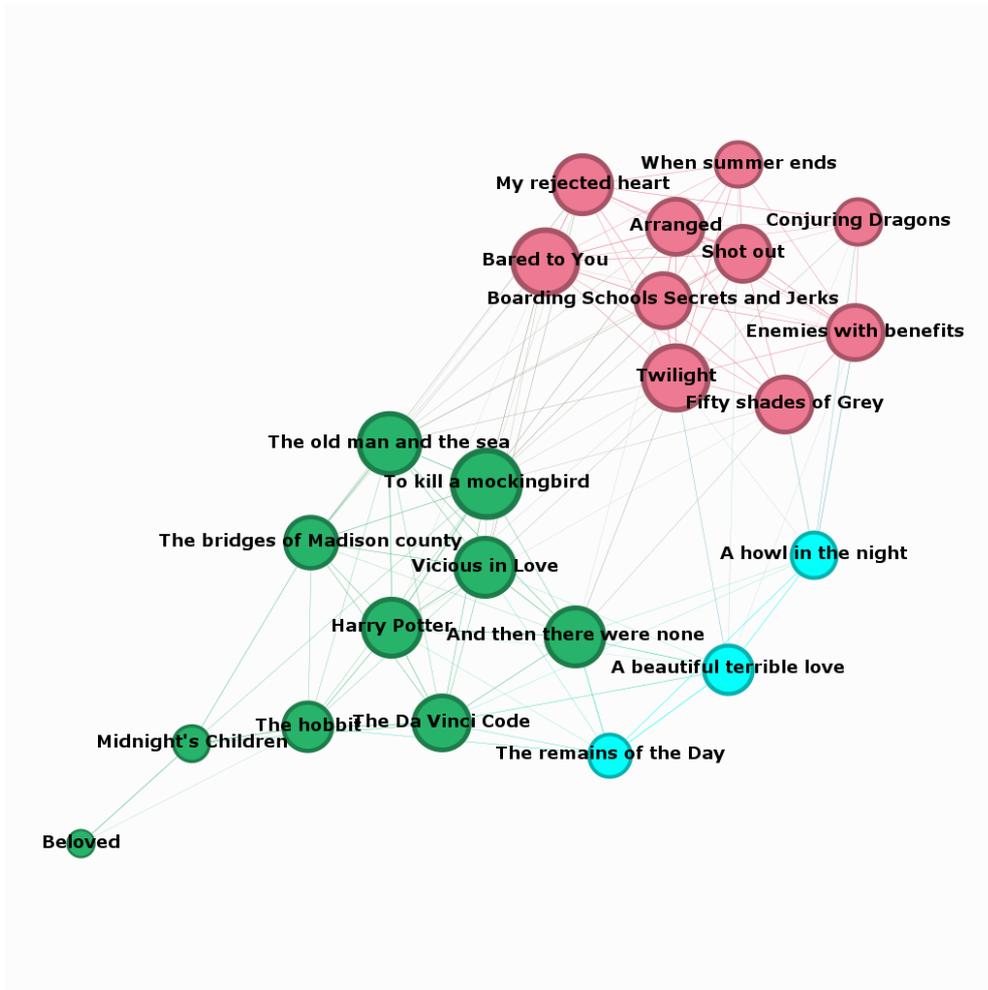

**Figure 1:** Generated Network of selected books

**Table 3**
Summary of the Classification

| Class 0 | Class 1 | Class 2 |
| --- | --- | --- |
| The Da Vinci Code | Arranged | **The remains of the Day*** |
| The hobbit | Shot out | A howl in the night |
| And then there were none | When summer ends | A beautiful, terrible love |
| The bridges of Madison county | Conjuring Dragons | |
| The old man and the sea | Enemies with benefits | |
| To kill a mockingbird | My rejected heart | |
| Midnight's Children | Twilight | |
| Harry Potter | Fifty shades of Grey | |
| Beloved | Bared to You | |
| **Vicious in Love*** | Boarding Schools, Secrets and Jerks | |

As shown in Table 3, the classes distinguish almost perfectly the Acclaimed Best-sellers. The only two anomalies are *The Remains of the Day* and *Vicious in Love*, which were assigned to unexpected classes.

---

[1] Awarded (Aw), High-Quality (HQ), Standard Quality (SQ) and Selph-Published (SP)
[2] See table 3 for the class composition
[3] POS = Aw+HQ; NEG = SQ+SP

It is also notable that the classification was unable to detect differences between Standard-Quality and Self-Published books.

### 4.1.1. Discussion

As previously noted, the algorithm effectively captures the distinctions between the two primary clusters of books. This represents a significant preliminary result, as the classification model operates without a training set and is therefore not influenced by a training phase. Furthermore, a semantic bias can be excluded; although semantic measures were included in the analysis, they were utilized strictly in a quantitative manner rather than for qualitative content interpretation. Finally, the four original classes contain romance novels (love stories) which were not automatically isolated into a separate category, suggesting that the model prioritizes stylistic and structural features over genre-specific thematic content.

The only classification errors involve the books *The Remains of the Day* and *Vicious in Love*. In terms of similarities, *The Remains of the Day* presents only two high values in the matrix, corresponding to the books "A Beautiful, Terrible Love" and "A Howl in the Night". In terms of single features, it presents many peculiarities: Its connective frequencies exhibit significant deviations from the average observed in other edited books, particularly concerning Additive Negative, Logical, and Temporal Negative connectives. Given the incidence of demonstrative articles — and notably the high frequency of the present tense (0.44) and first-person pronouns (0.32) shared with the two self-published books in its cluster—the classification error may be attributed to the narrative's first-person perspective, which blends synchronic descriptions with recollections. This constitutes a bias in the classification model, as the use of the first person is independent of narrative quality. An additional factor contributing to the segregation of these three ostensibly unrelated books into a distinct class may lie in their Lexical Ranges: while the overall group averages are comparable, the scores of these three specific volumes deviate from the mean in a consistent and similar manner.

In a future experiment, this metric can be avoided in order to prevent this kind of error.

The book *Vicious in Love* seems to be characterized by a more sophisticated language compared to other books in its class. It makes lower use of pronouns but presents the highest values of definite articles within the class. The lexicon is slightly more sophisticated, with the lowest score in Range-1 (0.72) contrasted by a high score in Range-3 (0.03).

In general, this qualitative analysis confirms the starting hypothesis: linguistic features are extremely useful to identify the quality of a book and allow for a good classification in almost all cases. Nevertheless, there are unpredictable factors which cannot be assessed solely by a linguistic analysis.

### 4.2. Evaluation of short narratives

The second experiment have been proposed to validate the effectiveness of extracted features on a large textual dataset, which allows a comparison among the proposed model and other traditional evaluation methods. Features were extracted from the HANNA dataset's 96 human-written texts and 959 texts equally generated by those models: BERTGeneration, CTRL, GPT, GPT-2(tag), GPT-2, RoBERTa, XLNet, Fusion, HINT and TD-VAE. All 1056 texts were evaluated by human judges, allowing for the calculation of a correlation between the proposed automatic metric and human ratings.

To generate a quality score using our proposed features, a similarity matrix was constructed for all 1056 texts, as described in section 3.3.2. The quality score was calculated as a Differential Polarity Index which takes into account the books of the Gold Standard subdivided in classes.

We tested four different classification and combination of average similarity scores obtaining the correlations illustrated in table 4

The optimal configuration was identified using the original classification while excluding the High-Quality (HQ) category from the core calculation. To evaluate the effectiveness of this approach, the results were benchmarked against established methodologies as presented by Chhun et al. [4]. The comparative analysis, based on Kendall correlation coefficients ($\tau$), is summarized in Table 5.

**Table 4**
Pearson and Kendall Correlations between Human ratings and the tested formulas. All the correlation are statistically significant (p-value < 0.05)

| Classification | Formula | Pearson | Kendall |
|---|---|---|---|
| Original | Aw+HQ-SQ-SP | 0.4397 | 0.3387 |
| | Aw-SP | 0.4853 | 0.3160 |
| | Aw-SP-SQ | 0.5237 | 0.34.50 |
| | Aw-SP+HQ | 0.1473 | 0.0987 |
| Automatic | C0-C1-C2 | 0.3318 | 0.2068 |
| | C0-C2 | 0.4049 | 0.2892 |
| | C0-C2+C1 | 0.1119 | 0.0767 |
| Merged | POS-NEG | 0.4249 | 0.3036 |

**Table 5**
Kendall Correlations between Human ratings and different automatic evaluation metrics

| DPI | Beluga-13B 1 | Llama-13B 1 | Mistral-7B 1 | ChatGPT 1 | BARTScore | BERTScore | BLEU | ROUGE-1 | ChrF |
|---|---|---|---|---|---|---|---|---|---|
| 34.5 | 25 | 16 | 20 | 18 | 6 | 16 | 12 | 18 | 18 |

Table 5 shows the Kendall correlation coefficients (multiplied by 100) for the proposed metric alongside a set of LLM-based, embedding-based, and classic metrics on the same dataset, as reported in Chhun et al. [4]. The p-values provide strong statistical evidence that the proposed metric correlates positively with human ratings. Notably, the use of linguistic features outperforms all other methods, even overcomes Beluga-13B, which achieve the best results with the same dataset, probably due to the influence of fine-tuning and the dimension of the model [4]. The significant correlations achieved by the proposed metric demonstrate that specific linguistic features serve as robust markers of narrative quality. This alignment with human perception confirms that computational stylistic analysis can effectively capture the nuances of literary value.

### 4.3. Discussion

The high performance achieved by a metric based on linguistic features was unexpected. Initially, the aim of collecting various stylistic metrics was to gather preliminary evidence that language significantly influences the evaluation of narrative quality. The second experiment was designed as a more rigorous test once the efficiency of the method was confirmed by the initial results.

The success of this second experiment is particularly noteworthy given that the narratives in the HANNA dataset are very brief, often consisting of only a few sentences. Nevertheless, the method was able to efficiently analyze the corpus and produce these results. Another confirmation of the method's robustness is that our "gold standard"—the 23 books from the first experiment—presents linguistic values that are entirely comparable to those of the shorter texts. Future research can further test this approach by expanding the gold standard with additional volumes.

The high correlations achieved through the extraction of linguistic features characterize the three proposed classifications. This suggests that the opposition between "masterpieces" and other novels is closely aligned with the criteria prioritized by human evaluators during judgment tasks. Higher correlation values correspond to formulas in which Awarded Books are contrasted with novels less appreciated by critics, suggesting that narratives selected for prestigious awards are almost universally recognized as qualitatively valid. On the contrary, the High-Quality (HQ) category has proven to be a confounding factor in the proposed metric.

Feature weighting requires specific consideration. As indicated by the findings of the initial experiment, several metrics failed to contribute effectively to the final score. This is particularly evident with first-person pronouns, which do not serve as reliable indicators of literary value. Correlation scores increased by several percentage points when the weight of these features was decreased or when they were removed entirely. Furthermore, lexical range indicators appeared too uniform across the corpus to represent a meaningful distinction between quality levels; consequently, the overall score improved when these three indicators were excluded from the analysis.

Other metrics yielded better results only when their influence was attenuated. The frequency of past tenses and relative clauses, for instance, does not appear to be directly implicated in quality evaluation. A separate consideration applies to connectives: their sparsity, especially in shorter texts where many values remain at zero, tends to make vectors artificially similar. While the complete removal of these metrics did not lead to an increase in correlation, assigning them a minor weight significantly improved the final performance of the model.

From a qualitative perspective, comparing human-written and AI-generated texts in the HANNA dataset reveals that humans generally produce longer texts. Key differences include:

- **Noun-Pronoun Ratio**: Humans utilize more nouns, whereas generative models show a preference for pronouns;
- **D-textual-Value**: Human-authored texts exhibit a significantly higher average value (196) compared to generated texts (92), which means more rich and complex texts;
- **Connectives**: While temporal connectives are more frequent in human texts, other connective types are more balanced; surprisingly, negative logical connectives appear with higher frequency in generated texts on average. This finding contrasts with the assumptions previously held by Maisto [10] and Johansson [11] and open to further investigations. However, in the shorter narratives of the HANNA dataset, these features are frequently absent or exhibit frequencies too low to be effectively incorporated into the final quality score. This sparsity limits their reliability as markers of quality in brief texts, where their absence may be a result of length constraints rather than a specific stylistic choice.

These differences are generally unrelated to the overall narrative quality, as human subjects did not distinguish between them in terms of assigned quality scores; furthermore, automated evaluations revealed no significant distinctions.

## 5. Conclusion

This paper focused on linguistic features as effective tools for automatic narrative evaluation. Generally, this task is performed using techniques designed for Machine Translation Evaluation (MTE) or, more recently, through Embeddings and Large Language Models (LLMs). However, traditional methods often lack the precision of human evaluators, who remain the "gold standard" due to the diverse features they consider, such as character plausibility, plot, interestingness, and emotional impact. As demonstrated in this study, even a narrative that excels in these areas may fail if the writing quality is poor. Thus, while the linguistic dimension alone cannot fully evaluate a story, it must be considered a crucial component of the process, particularly in automated settings.

We proposed a set of 33 quantitative linguistic measures spanning lexical diversity, syntactic complexity, coherence, and semantic creativity. Our model includes established metrics such as D-Textual-Value and Lexical Range, a detailed syntactic analysis of sentence structures—with a specific focus on verb tenses—and the Textual Figurative Index, an original methodology designed to automatically quantify figurative language.

In the first experiment, the model analyzed a corpus of 23 books categorized into four classes: Awarded Books (Aw), High-Quality Best Sellers (HQ), Standard Quality Best Sellers (SQ), and Self-published Books (SP). The results, illustrated in Figure 1, demonstrate that the selected features effectively distinguish between acclaimed and criticized works. While the model faced challenges in perfectly differentiating between the four specific subclasses, it achieved high internal coherence.

The second experiment yielded even more significant results, with the proposed method outperforming all traditional evaluation metrics. These findings confirm that a stylistic analysis—relying on rule-based resources supported by BERT and CoreNLP—can be more effective than LLMs in tasks typically considered difficult for non-human evaluators. Indeed, relying on LLMs to judge other LLM-generated texts risks becoming a vicious circle where the evaluator and the evaluated share the same inherent biases.

Future work will expand this methodology to a larger corpus and incorporate additional features, such as emotion analysis and direct speech extraction. Furthermore, we intend to integrate this approach with other disciplines, such as semiotics for plot and character analysis, and knowledge extraction for the automatic classification of narrative functions. The ultimate goal is to develop a robust evaluation framework that can not only assess existing narratives but also define the linguistic standards required for generative models to produce effective and high-quality stories.

## Declaration on Generative AI

During the preparation of this work, the author used Gemini-3 in order to: Grammar and spelling check. After using this tool, the author reviewed and edited the content as needed and takes full responsibility for the publication's content.